\documentclass{article} % For LaTeX2e
\usepackage{nips14submit_e,times}
\usepackage{hyperref}
\usepackage{url}

\usepackage{amsmath,bm}
\usepackage{amssymb}
\usepackage{algorithm}
\usepackage{algorithmic}
\usepackage{graphicx}
\usepackage{tabularx}
\usepackage{array,arydshln}
\usepackage{subcaption}
\title{{\fontsize{15}{16}\selectfont Explain Images with Multimodal Recurrent Neural Networks}}

\author{
Junhua Mao$^{1,2}$\ \ \ \ \ \ \ \ Wei Xu$^1$ \ \ \ \ \ \ \ \ Yi Yang$^1$ \ \ \ \ \ \ \ \ Jiang Wang$^1$ \ \ \ \ \ \ \ \ Alan L. Yuille$^2$\\
$^1$Baidu Research \ \ \ \ \ \ \ $^2$University of California, Los Angeles\\
{\fontsize{8}{9}\selectfont \texttt{mjhustc@ucla.edu}, \texttt{\{wei.xu,yangyi05,wangjiang03\}@baidu.com},  \texttt{yuille@stat.ucla.edu}}
}

\nipsfinalcopy % Uncomment for camera-ready version

\begin{document}

\maketitle

\vspace{-1em}
\begin{abstract}

In this paper, we present a multimodal Recurrent Neural Network (m-RNN) model for generating novel sentence descriptions to explain the content of images.
It directly models the probability distribution of generating a word given previous words and the image.
Image descriptions are generated by sampling from this distribution.
The model consists of two sub-networks: a deep recurrent neural network for sentences and a deep convolutional network for images. 
These two sub-networks interact with each other in a multimodal layer to form the whole m-RNN model.
The effectiveness of our model is validated on three benchmark datasets (IAPR TC-12 \cite{grubinger2006iapr}, Flickr 8K \cite{rashtchian2010collecting}, and Flickr 30K \cite{hodoshimage}).
Our model outperforms the state-of-the-art generative method.
In addition, the m-RNN model can be applied to retrieval tasks for retrieving images or sentences, and achieves significant performance improvement over the state-of-the-art methods which directly optimize the ranking objective function for retrieval.

\end{abstract}

\section{Introduction}

Obtaining sentence level descriptions for images is becoming an important task and has many applications, such as early childhood education, image retrieval, and navigation for the blind.
Thanks to the rapid development of computer vision and natural language processing technologies, recent works have made significant progress for this task (see a brief review in Section \ref{sec:related_work}).
Many of these works treat it as a retrieval task.
They extract features for both sentences and images, and map them to the same semantic embedding space.
These methods address the tasks of retrieving the sentences given the query image or retrieving the images given the query sentences.
But they can only label the query image with the sentence annotations of the images already existing in the datasets, thus lack the ability to describe new images that contain previously unseen combinations of objects and scenes.

In this work, we propose a multimodal Recurrent Neural Networks (m-RNN) model to address both the task of generating novel sentences descriptions for images, and the task of image and sentence retrieval.
The whole m-RNN architecture contains a language model part, an image part and a multimodal part.
The language model part learns the dense feature embedding for each word in the dictionary and stores the semantic temporal context in recurrent layers.
The image part contains a deep Convulutional Neural Network (CNN) \cite{krizhevsky2012imagenet} which extracts image features.
The multimodal part connects the language model and the deep CNN together by a one-layer representation.
Our m-RNN model is learned using a perplexity based cost function (see details in Section \ref{sec:trainCost}).
The errors are backpropagated to the three parts of the m-RNN model to update the model parameters simultaneously.
To the best of our knowledge, this is the first work that incorporates the Recurrent Neural Network in a deep multimodal architecture.

In the experiments, we validate our model on three benchmark datasets: IAPR TC-12 \cite{grubinger2006iapr}, Flickr 8K \cite{rashtchian2010collecting}, and Flickr 30K \cite{hodoshimage}.
we show that our method significantly outperforms the state-of-the-art methods in both the task of generating sentences and the task of image and sentence retrieval when using the same image feature extraction networks.
Our model is extendable and has the potential to be further improved by incorporating more powerful deep networks for the image and the sentence.

\begin{figure}[tb!]
\begin{center}
\includegraphics[width=0.98\linewidth]{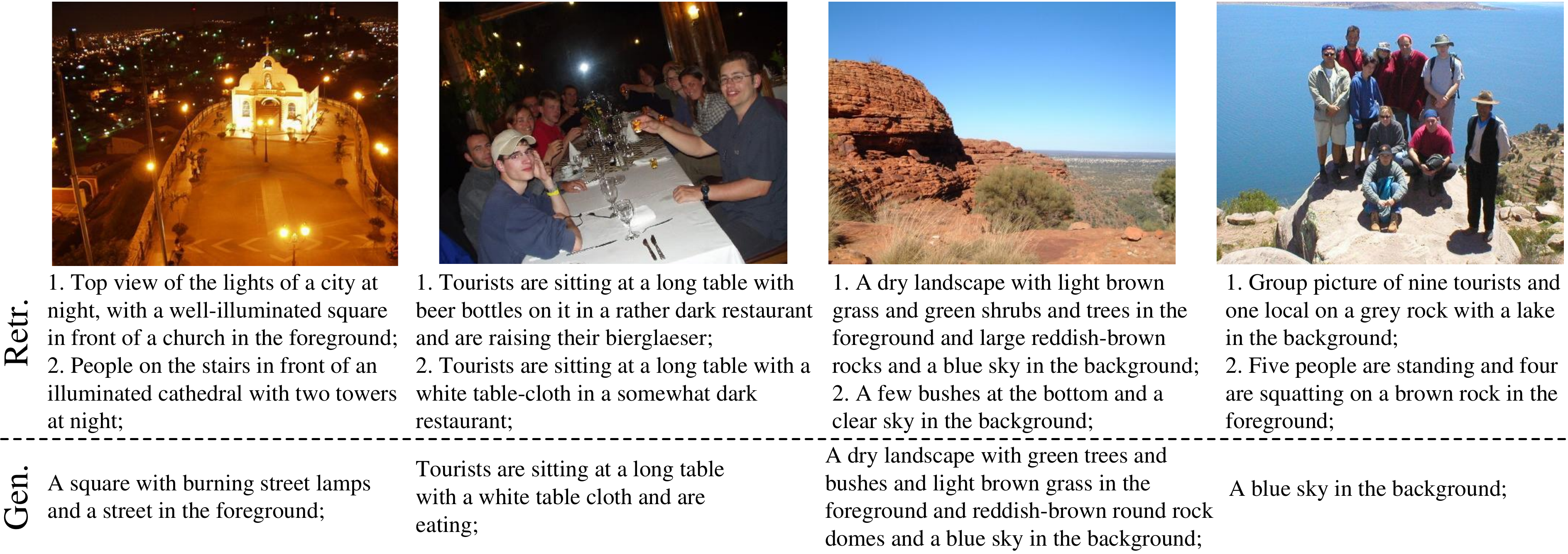}
\end{center}
   \caption{Examples of the generated and two top-ranked retrieved sentences given the query image from IAPR TC-12 dataset.
   The sentences can well describe the content of the images.
   We show a failure case in the fourth image, where the model mistakenly treats the lake as the sky.
   }
\label{fig:res_example}
\end{figure}

\section{Related Work}
\label{sec:related_work}

\textbf{Deep model for computer vision and natural language.}
The deep neural network structure develops rapidly in recent years in both the field of computer vision and natural language.
For computer vision, Krizhevsky et. al \cite{krizhevsky2012imagenet} proposed a deep convolutional neural networks with 8 layers (denoted as AlexNet) for image classification tasks and outperformed previous methods by a large margin.
Recently, Girshick et. al \cite{girshick2014rcnn} proposed a object detection framework based on AlexNet.
For natural language, the Recurrent Neural Network shows the state-of-the-art performance in many tasks, such as speech recognition and word embedding learning \cite{mikolov2010recurrent,mikolov2011extensions,mikolov2013distributed}.

\textbf{Image-sentence retrieval.}
Many works treat the task of describe images as a retrieval task and formulate the problem as a ranking or embedding learning problem \cite{hodosh2013framing,frome2013devise,socher2014grounded}.
They will first extract the word and sentence features (e.g. Socher et.al \cite{socher2014grounded} uses dependency tree Recursive Neural network to extract sentence features) as well as the image features.
Then they optimize a ranking cost to learn an embedding model that maps both the language feature and the image feature to a common semantic feature space.
In this way, they can directly calculate the distance between images and sentences.
Most recently, Karpathy et.al \cite{karpathy2014fragment} showed that object level image features based on object detection results will generate better results than image features extracted at the global level.

\textbf{Generating novel sentence descriptions for images.}
There are generally two categories of methods for this task.
The first category assumes a specific rule of the language grammar.
They parse the sentence and divide it into several parts \cite{mitchell2012midge,gupta2012image}.
Then each part is associated to a object or an attribute in the image (e.g. \cite{kulkarni2011baby} uses a Conditional Random Field model and \cite{farhadi2010every} uses a Markov Random Field model).
This kind of method generates sentences that are syntactically correct.
Another category of methods, which is more related to our method, learns a probability density over the space of multimodal inputs (i.e. sentences and images), using for example, Deep Boltzmann Machines \cite{srivastava2012multimodal}, and topic models \cite{barnard2003matching,jia2011learning}.
They can generate sentences with richer and more flexible structure than the first group.
The probability of generating sentences given the corresponding image can serves as the affinity metric for retrieval.
Our method falls into this category.
More close related to our tasks and method is the work of Kiros et al. \cite{kiros2013multimodal}, which is built on a Log-BiLinear model \cite{mnih2007three}.
It needs a fixed length of context (i.e. five words), whereas in our model, the temporal context is stored in a recurrent architecture, which allows arbitrary context length.

\section{Model Architecture}

\begin{figure}[htb]
\begin{center}
\includegraphics[width=0.95\linewidth]{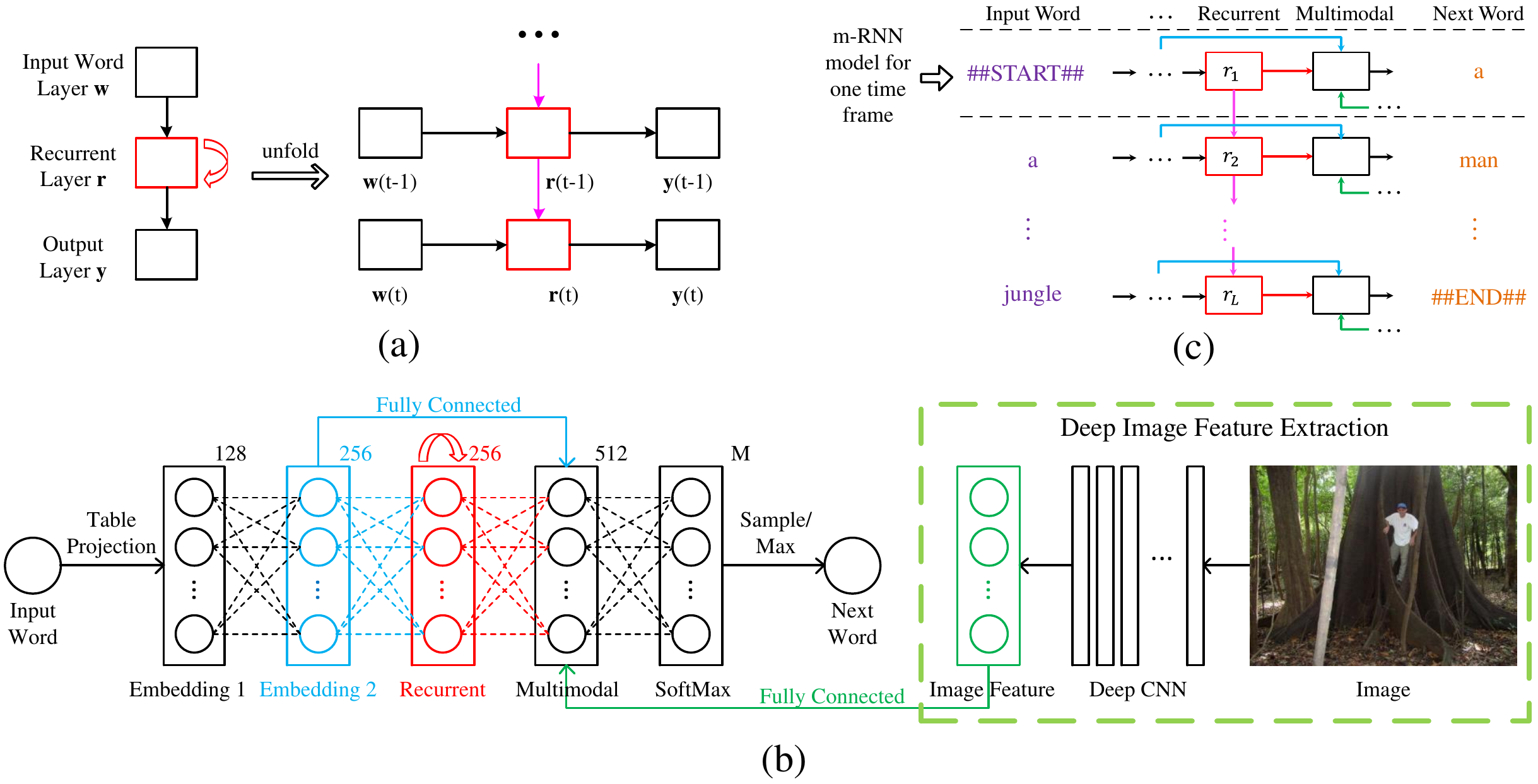}
\end{center}
   \caption{Illustration of the simple Recurrent Neural Network (RNN) and our multimodal Recurrent Neural Network (m-RNN) architecture.
   (a). The simple RNN. 
   (b). Our m-RNN model.
   The input of our model is an image and its corresponding sentences (e.g. the sentence for the shown image is: \emph{a man at a giant tree in the jungle}).
   The model will estimate the probability distribution of the next word given previous words and the image.
   This architecture is much deeper than the simple RNN.
   % of structure without considering the extendibility of the recurrent layer.
   (c). The illustration the unfolded m-RNN. 
   % We can unfold the recurrent layer, which leads to the temporal depth of the network.
   The model parameters are shared for each temporal frame of the m-RNN model.
   }
\label{fig:illu_RNN}
\end{figure}

% For layer \textcircled{\raisebox{-0.9pt}{1}} and layer \textcircled{2},
\subsection{Simple recurrent neural network}
\label{sec:sRNN}
We briefly introduce the simple Recurrent Neural Network (RNN) or Elman network \cite{elman1990finding} that is widely used for many natural language processing tasks, such as speech recognition \cite{mikolov2010recurrent,mikolov2011extensions}.
Its architecture is shown in Figure \ref{fig:illu_RNN}(a).
It has three types of layers in each time frame: the input word layer $\mathbf{w}$, the recurrent layer $\mathbf{r}$ and the output layer $\mathbf{y}$.
The activation of input, recurrent and output layers at time $t$ is denoted as $\mathbf{w}(t)$, $\mathbf{r}(t)$, and $\mathbf{y}(t)$ respectively.
$\mathbf{w}(t)$ is the one-hot representation of the current word. 
This representation is binary, and has the same dimension of the vocabulary size with only one non-zero element.
$y(t)$ can be calculated as follows:
\begin{equation}
\mathbf{x}(t) = [\mathbf{w}(t)\ \ \mathbf{r}(t-1)];\ \ \ 
\mathbf{r}(t)=f_1(\mathbf{U} \cdot \mathbf{x}(t));\ \ \ 
\mathbf{y}(t)=g_1(\mathbf{V} \cdot \mathbf{r}(t));
\end{equation}
where $\mathbf{x}(t)$ as a vector that concatenates $\mathbf{w}(t)$ and $\mathbf{r}(t-1)$, $f_1(.)$ and $g_1(.)$ are element-wised sigmoid and softmax function respectively, and $\mathbf{U}$, $\mathbf{V}$ are weights which will be learned.

The size of RNN is adaptive to the length of the input sequence and the recurrent layers connect the sub-networks in different time frames.
Accordingly, when we do the backpropagation, we need to propagate the error through recurrent connections back in time \cite{rumelhart1988learning}.

\subsection{Our m-RNN model}
The structure of our multimodal Recurrent Neural Network (m-RNN) is shown in Figure \ref{fig:illu_RNN}(b).
The m-RNN model is much deeper than the simple RNN model.
It has six layers in each time frame: the input word layer, two word embedding layers, the recurrent layer, the multimodal layer, and the softmax layer).

The two word embedding layers embed the one-hot input into a dense word representation.
It has several advantages.
Firstly, it will significantly lower the number of parameters in the networks because the dense word vector (128 dimension) is much smaller than the one-hot word vector.
Secondly, the dense word embedding encodes the semantic meanings of the words \cite{mikolov2013efficient}.
The semantically relevant words can be found by calculating the Euclidean distance between two dense word vectors in embedding layers.

Most of the sentence-image multimodal models \cite{karpathy2014fragment,frome2013devise,socher2014grounded,kiros2013multimodal} use pre-computed word embedding vectors as the initialization of their model. 
In contrast, we randomly initialize our word embedding layers and learn them from the training data.
We show that this random initialization is sufficient for our architecture to generate the state-of-the-art results.
% To further refine the word representation, we add a hidden layer after the initial word embedding layer and 
We treat the activation of the word embedding layer 2 (see Figure \ref{fig:illu_RNN}(b)) as the final word representation, which directly inputs in the multimodal layer.

After the two word embedding layers, we have a recurrent layer with 256 dimensions.
The calculation of the recurrent layer is slightly different from the calculation for the simple RNN.
Instead of concatenating the word representation at time $t$ (denoted as $\mathbf{w}(t)$) and the recurrent layer activation at time $t-1$ (denoted as $\mathbf{r}(t-1)$), we first map $\mathbf{r}(t-1)$ into the same vector space as $\mathbf{w}(t)$ and add them together:
\begin{equation}
\mathbf{r}(t)=f_2(\mathbf{U}_r \cdot \mathbf{r}(t-1) + \mathbf{w}(t));
\end{equation}
% where $\mathbf{s}$ and $\mathbf{w}$ denotes the recurrent layer vector and the word representation respectively.
% This strategy will reduce the number of parameters and accelerate the training and testing process.
We set $f_2(.)$ as the Rectified Linear Unit (ReLU), inspired by its the recent success when training very deep structure in computer vision field \cite{krizhevsky2012imagenet}.
This differs from the simple RNN where the sigmoid function is adopted (see Section \ref{sec:sRNN}).
ReLU is faster, and harder to saturate or overfit the data than non-linear functions like the sigmoid.
When backpropagation through time (BPTT) \cite{rumelhart1988learning} is conducted for RNN with sigmoid function, the vanishing gradient problem appears since even the simplest RNN model can have a large temporal depth.
Previous methods \cite{mikolov2010recurrent,mikolov2011extensions} used heuristics, such as truncated BPTT, to avoid this problem.
Truncated BPTT stops the BPTT after $k$ time steps, where $k$ is a hand-defined hyperparameter.
Because of the good properties of ReLU, we do not need to stop the BPTT at an early stage, which leads to a better and more efficient utilization of the data than truncated BPTT.

After the recurrent layer, we set up a 512 dimensional multimodal layer that connect the language model part and the image part of the m-RNN model (see Figure \ref{fig:illu_RNN}(b)).
The language model part includes the word embedding layer 2 (the final word representation) and the recurrent layer (the sentence context).
The image part contains the image feature extraction network.
Here we connect the seventh layer of AlexNet \cite{krizhevsky2012imagenet} to the multimodal layer (please refer to Section \ref{sec:ImgSenFeat} for more details).
But our framework can use any image features.
We map the feature vector for each layer to the same feature space and add them together to obtain the feature vector for the multimodal layer:
\begin{equation}
\mathbf{m}(t)=g_2(\mathbf{V}_w \cdot \mathbf{w}(t) + \mathbf{V}_r \cdot \mathbf{r}(t) + \mathbf{V}_I \cdot \mathbf{I});
\end{equation}
where $\mathbf{m}$ denotes the multimodal layer feature vector, $\mathbf{I}$ denotes the image feature, $g_2(.)$ is the element-wised scaled hyperbolic tangent function \cite{lecun2012efficient}:
\begin{equation}
g_2(x) = 1.7159 \cdot \tanh( \frac{2}{3} x)
\end{equation}
This function forces the gradients into the most non-linear value range and accelerates the training process than the basic hyperbolic tangent function.

%We do not restrict the norm the three feature layers that connect to the multimodal layer.
%Our experiments shows that the L2 norm of the hidden layer, 

As the simple RNN, our m-RNN model has a softmax layer that will generate the probability distribution of the next word.
The dimension of this layer is the vocabulary size $M$, which is different for different datasets.

\section{Training the m-RNN}
\label{sec:trainCost}
For training our m-RNN model we adopt a cost function based on the \emph{Perplexity} of the sentences in the training set given their corresponding images.
Perplexity is a standard measure for evaluating language model.
The perplexity for one word sequence (i.e. a sentences) $w_{1:L}$ is calculated as follows:
\begin{equation}
\log_2 \mathcal{PPL}(w_{1:L}|\mathbf{I}) = -\frac{1}{L} \sum_{n=1}^{L} \log_2 P(w_n|w_{1:n-1},\mathbf{I})
\end{equation}
where $L$ is the length of the word sequences, $\mathcal{PPL}(w_{1:L}|\mathbf{I})$ denotes the perplexity of the sentence $w_{1:L}$ given the image $\mathbf{I}$.
$P(w_n|w_{1:n-1},\mathbf{I})$ is the probability of generating the word $w_n$ given $\mathbf{I}$ and previous words $w_{1:n-1}$.
It corresponds to the feature vector of the SoftMax layer of our model.

The cost function of our model is the average log-likelihood of the words given their context words and corresponding images in the training sentences plus a regularization term.
It can be calculated by the perplexity:
\begin{equation}
\mathcal{C} = \frac{1}{N} \sum_{i=1}^{N} L \cdot \log_2 \mathcal{PPL}(w_{1:L}^{(i)}|\mathbf{I}^{(i)}) + \left \| \theta \right \|_2^2
\end{equation}
where $N$ is the number of words in the training set and $\theta$ is the model parameters.
% It is equivalent to the reciprocal of the geometric mean of the probability to generate the training sentences using the model.

Our training objective is to minimize this cost function, which is equivalent to maximize the probability of the model to generate the sentences in the training set given their corresponding images.
The cost function is differentiable and we use backpropagation to learn the model parameters.

\section{Learning of Sentence and Image Features}
\label{sec:ImgSenFeat}

The architecture of our model allows the gradients from the loss function to be backpropagated to both the language modeling part (i.e. the word embedding layers and the recurrent layer) as well as the image part (e.g. the AlexNet \cite{krizhevsky2012imagenet}).

For the language modeling part, as mentioned above, we randomly initialize the language modeling layers and learn their parameters. For the image part, we connect the seventh layer of a pre-trained Convolutional Neural Network \cite{krizhevsky2012imagenet,donahue2013decaf} (denoted as AlexNet).
The same features extracted from the seventh layer of AlexNet (also denoted as decaf features \cite{donahue2013decaf}) are widely used by previous multimodal methods \cite{kiros2013multimodal,frome2013devise,karpathy2014fragment,socher2014grounded}.
A recent multimodal retrieval work \cite{karpathy2014fragment} showed that using the RCNN object detection framework \cite{girshick2014rcnn} combined with the decaf features significantly improves the performance.
In the experiments, we show that our method performs much better than \cite{karpathy2014fragment} when the same image features are used, and is better than or comparable to their results even when they use more sophisticated features based on object detection.

We can update the AlexNet according to the gradient backpropagated from the multimodal layer. In this paper, we fix the image features and the deep CNN network in the training stage due to a shortage of data (The datasets we used in the experiment have less than 30K images).
In future work, we will apply our method on large datasets and finetune the parameters of the deep CNN network in the training stage.

\section{Sentence Generation, Image and Sentence Retrieval}
We can use the trained m-RNN model for three tasks: 1) Sentences generation; 2) Sentence retrieval (retrieving most relevant sentences to the given image); 3) Image retrieval (retrieving most relevant images to the given sentence);

The sentence generation process is straightforward.
Start from the start sign ``\#\#START\#\#'' or arbitrary number of reference words (e.g. we can input the first K words in the reference sentence to the model and then start to generate new words), our model can calculate the probability distribution of the next word: $P(w|w_{1:n-1},\mathbf{I})$.
Then we can sample from this probability distribution to pick the next word.
In practice, we find that selecting the word with the maximum probability performs slightly better than sampling.
After that, we input the picked word to the model and continue the process until the model outputs the end sign ``\#\#END\#\#''.

For the retrieval tasks, we use our model to calculate the perplexity of generating a sentence given an image.
The perplexity can be treated as an affinity measurement between sentences and images.
For the image retrieval task, we rank the images based on their perplexity with the query sentence and output the top ranked ones. 

The sentence retrieval task is trickier because there might be some sentences that have high probability for any image query (e.g. sentences consists of many frequently appeared words).
Instead of looking at the perplexity or the probability of generating the sentences given the query image, we use the normalized probability for each sentence: $P(w_{1:L}|\mathbf{I}) / P(w_{1:L})$.
$P(w_{1:L}) = \sum_{\mathbf{I^{'}}} P(w_{1:L}|\mathbf{I^{'}}) \cdot P(\mathbf{I^{'}})$, where $\mathbf{I^{'}}$ are images sampled from the training set.
We approximate $P(\mathbf{I^{'}})$ by a constant and ignore this term.
$P(w_{1:L}|\mathbf{I}) = \mathcal{PPL}(w_{1:L}|\mathbf{I}) ^ {-L}$.

\section{Experiments}

\subsection{Datasets}
We test our method on three benchmark datasets with sentence level annotations: IAPR TC-12 \cite{grubinger2006iapr}, Flickr 8K \cite{rashtchian2010collecting}, and Flickr 30K \cite{hodoshimage}.

% Here are some statistics and our experimental settings for the three datasets:

\textbf{IAPR TC-12 Benchmark}.
This dataset consists of around 20,000 images taken from locations around the world.
This includes images of different sports and actions, people, animals, cities, landscapes, and so on.
For each image, they provide at least one sentence annotation.
On average, there are about 1.7 sentences annotations for one image.
We adopt the publicly available separation of training and testing set as previous works \cite{GVS10a,kiros2013multimodal}.
There are 17,665 images for training and 1962 images for testing.

\textbf{Flickr8K Benchmark}.
This dataset consists of 8,000 images extracted from Flickr.
For each image, it provides five sentences annotations.
The grammar of the annotations for this dataset is simpler than that for the IAPR TC-12 dataset.
We adopt the standard separation of training, validation and testing set which is provided by the dataset.
There are 6,000 images for training, 1,000 images for validation and 1,000 images for testing.

\textbf{Flickr30K Benchmark}.
This dataset is a recent extension of Flickr8K.
For each image, it also provides five sentences annotations.
It consists of 158,915 crowd-sourced captions describing 31,783 images.
The grammar and style for the annotations of this dataset is similar to Flickr8K.
We follow the previous work \cite{karpathy2014fragment} which used 1,000 images for testing.
This dataset, as well as the Flick8K dataset, are commonly used for the image-sentence retrieval tasks.
%and there is not public available results of methods for generating novel sentence descriptions.

\subsection{Evaluation metrics}
\label{sec:EvaRet}

\textbf{Sentence Generation}.
Following previous works, we adopt sentence perplexity and BLEU scores (i.e. B-1, B-2, and B-3) \cite{papineni2002bleu,lin2004automatic} as the evaluation metrics.
BLEU scores were originally designed for automatic machine translation where they rate the quality of a translated sentences given several references sentences.
We can treat the sentence generation task as the ``translation'' of the content of images to sentences.
BLEU remains the standard evaluation metric for sentence generation methods for images, though it has drawbacks.
For some images, the reference sentences might not contains all the possible descriptions in the image and BLEU might penalize some correctly generated sentences. 
To conduct a fair comparison, we adopt the same sentence generation steps and experiment settings as \cite{kiros2013multimodal}, and generate as many words as there are in the reference sentences to calculate BLEU.
Note that our model does not need to know the length of the reference sentence because we add a end sign "\#\#END\#\#" at the end of every training sentences and we can stop the generation process when our model outputs the end sign.

\textbf{Sentence Retrieval and Image Retrieval}
For Flickr8K and Flickr30K datasets, we adopted the same evaluation metrics as previous works \cite{socher2014grounded,frome2013devise,karpathy2014fragment} for both the tasks of sentences retrieval and image retrieval.
They used R@K (K = 1, 5, 10) as the measurements, which are the recall rates of the first retrieved groundtruth sentences (sentence retrieval task) or images (image retrieval task).
Higher R@K usually mean better retrieval performance.
Since we care most about the top-ranked retrieved results, the R@K with small K are more important.
The Med r is another score we used, which is the median rank of the first retrieved groundtruth sentences or images.
Lower Med r usually means better performance.

For IAPR TC-12 datasets, we adopt exactly the same evaluation metrics as \cite{kiros2013multimodal} and plot the mean number of matches of the retrieved groundtruth sentences or images with respect to the percentage of the retrieved sentences or images for the testing set.
For sentences retrieval task, \cite{kiros2013multimodal} used a shortlist of 100 images which are the nearest neighbors of the query image in the feature space.
This shortlist strategy makes the task harder because similar images might have similar descriptions and it is often harder to find  subtle differences among the sentences and pick the most suitable one.
Although there are no published R@K scores and Med r score for this dataset available for the best of our knowledge, we also report these scores of our method for future comparison.

\subsection{Results on IAPR TC-12}

\begin{table}[htb]
	\centering
\begin{tabular}{l|cccc}
\hline
      & $\mathcal{PPL}$  & B-1   & B-2   & B-3 \\
\hline
BACK-OFF GT2 & 54.5  & 0.323 & 0.145 & 0.059 \\
BACK-OFF GT3 & 55.6  & 0.312 & 0.131 & 0.059 \\
LBL \cite{mnih2007three}  & 20.1  & 0.327 & 0.144 & 0.068 \\
MLBL-B-DeCAF \cite{kiros2013multimodal} & 24.7  & 0.373 & \textbf{0.187} & 0.098 \\
MLBL-F-DeCAF \cite{kiros2013multimodal} & 21.8  & 0.361 & 0.176 & 0.092 \\
Gupta et al. \cite{gupta2012choosing} & /     & 0.15  & 0.06  & 0.01 \\
Gupta \& Mannem \cite{gupta2012image} & /     & 0.33  & 0.18  & 0.07 \\
\hdashline
Ours-RNN-Base & 7.77  & 0.3134 & 0.1168 & 0.0803 \\
Ours-m-RNN & \textbf{6.92} & \textbf{0.3951} & 0.1828 & \textbf{0.1311} \\
\hline
\end{tabular}%
	\caption{Results of the sentence generation task on the IAPR TC-12 dataset. ``B'' is short for BLUE}
	\label{tab:iaprtc_gen}
\end{table}

The results of the sentence generation task are shown in Table \ref{tab:iaprtc_gen}.
BACK-OFF GT2 and GT3 are n-grams methods with Katz backoff and Good-Turing discounting \cite{chen2000survey,kiros2013multimodal}.
Ours-RNN-Base serves as a baseline method for our m-RNN model.
It has the same architecture with m-RNN except that we will not input the image features to the network.

To conduct a fair comparison, we followed the same experimental settings of \cite{kiros2013multimodal}, include the context length to calculate the BLEU scores and perplexity.
These two evaluation metrics are not necessarily correlated to each other for the following reasons.
As mentioned in Section \ref{sec:trainCost}, perplexity is calculated according to the conditional probability of the word in a sentence given all of its previous reference words.
Therefore, a strong language model that successfully captures the distributions of words in sentences can have a low perplexity without the image content.
But the content of the generated sentences might be unrelated to images.
From Table \ref{tab:iaprtc_gen}, we can see that although our baseline method of RNN generates a very low perplexity, its BLEU score is not very high, indicating that it failed to generate sentences with high quality.

We show that our m-RNN model performs much better than our baseline RNN model in terms of both perplexity and BLEU score.
It also outperforms the state-of-the-art methods in terms of perplexity, B-1, B-3, and a comparable result for B-2 \footnote{\cite{kiros2013multimodal} further improve their results after the publication. The perplexity of MLBL-F and LBL now are 9.90 and 9.29 respectively.}.

\begin{figure}[htb]
        \centering
        \begin{subfigure}[b]{0.42\textwidth}
                \includegraphics[width=\textwidth]{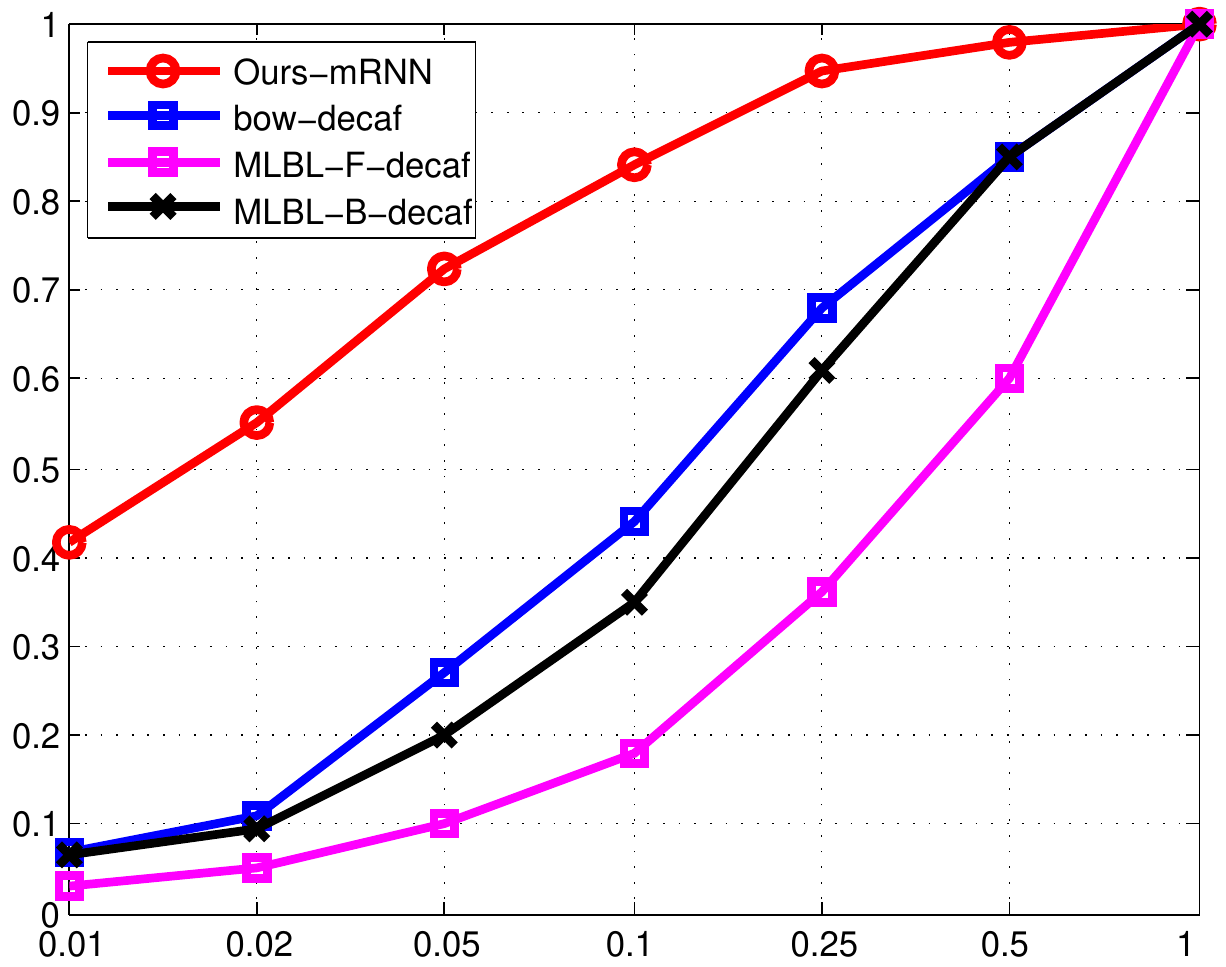}
                \caption{Image to Text Curve}
        \end{subfigure}%
        ~ %add desired spacing between images, e. g. ~, \quad, \qquad, \hfill etc.
          %(or a blank line to force the subfigure onto a new line)
        \begin{subfigure}[b]{0.42\textwidth}
                \includegraphics[width=\textwidth]{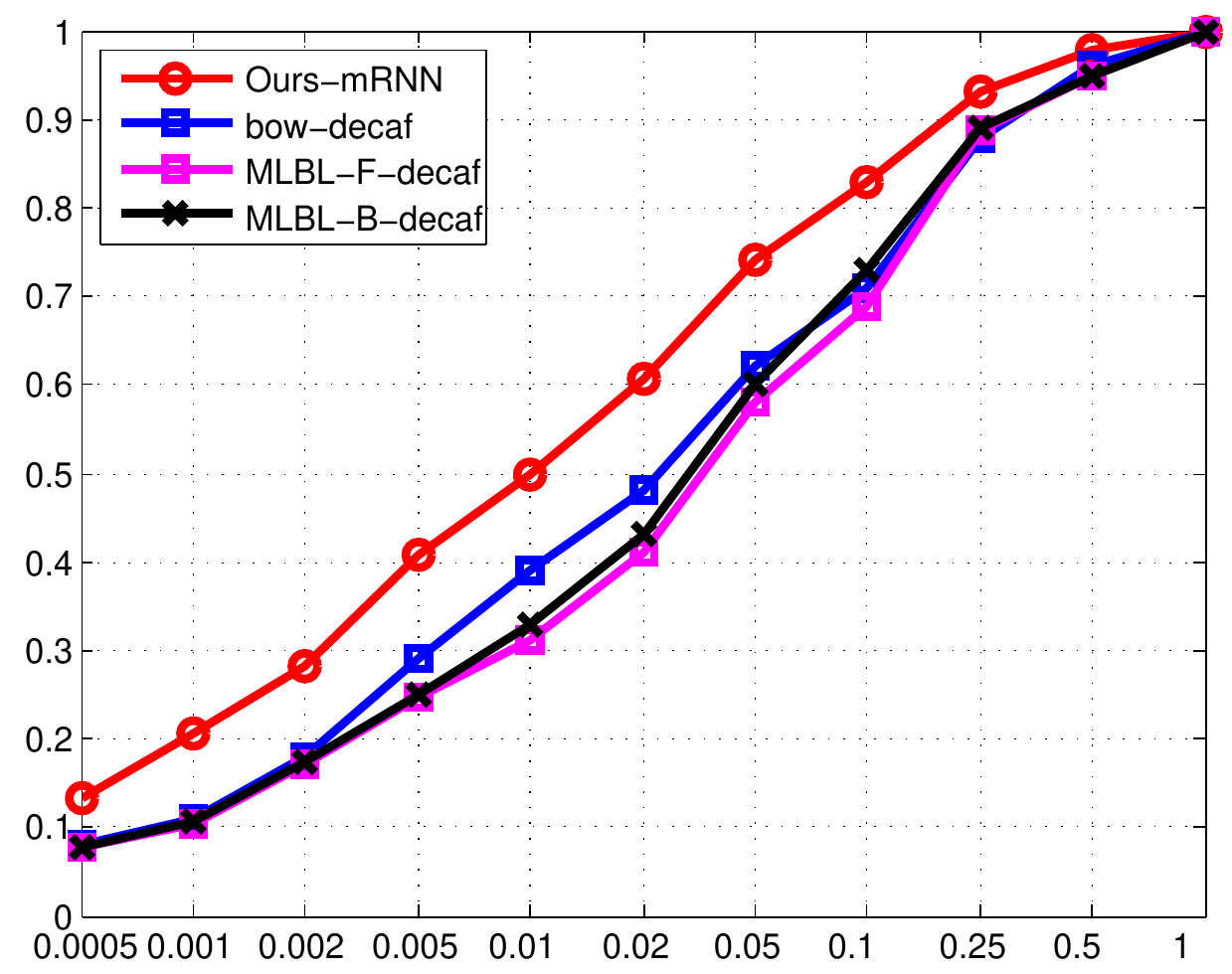}
                \caption{Text to Image Curve}
        \end{subfigure}
        \caption{Retrieval recall curve for (a). Sentence retrieval task (b). Image retrieval task on IAPR TC-12 dataset. The behavior on the far left (i.e. top few retrievals) is most important.}
        \label{fig:iaprtc_ret_curve}
\end{figure}

For retrieval tasks, as mentioned in Section \ref{sec:EvaRet}, we draw a recall accuracy curve with respect to the percentage of retrieved images (sentence retrieval task) or sentences (sentence retrieval task) in Figure \ref{fig:iaprtc_ret_curve}.
For sentence retrieval task, we used a shortlist of 100 images as the three comparing methods shown in \cite{kiros2013multimodal}.
The first method, bow−decaf, is a strong image based bag-of-words baseline.
The second and the third models are all multimodal deep models.
Our m-RNN model significantly outperforms these three methods in this task.

Since there are no publicly available results of R@K and median rank in this dataset, we report R@K scores of our method in Table \ref{tab:iaprtc_ret} for future comparisons.
The result shows that 20.9\% top-ranked retrieved images and 13.2\% top-ranked retrieved sentences are groundtruth.

\begin{table}[htb]
	\centering
\begin{tabular}{l|cccc|cccc}
\hline
      & \multicolumn{4}{c|}{Sentence Retrival (Image to Text)} & \multicolumn{4}{c}{Image Retrival (Text to Image)} \\
\hline
      & R@1   & R@5   & R@10  & Med r & R@1   & R@5   & R@10  & Med r \\
\hline
Ours-m-RNN & 20.9  & 43.8  & 54.4  & 8     & 13.2  & 31.2  & 40.8  & 21 \\
\hline
\end{tabular}%
	\caption{R@K and median rank (Med r) for iaprtc-12 dataset.}
	\label{tab:iaprtc_ret}
\end{table}

\subsection{Results on flickr8K}

This dataset was widely used as a benchmark dataset of image and sentence retrieval.
The R@K and Med r of different methods are shown in Table \ref{tab:flickr8K_ret}.
Our model outperforms the state-of-the-art methods (i.e Socher-decaf, DeViSE-decaf, DeepFE-decaf) by a large margin when using the same image features (i.e. decaf features).
We also list the performance of methods using more sophisticated features in Table \ref{tab:flickr8K_ret}.
``-avg-rcnn'' denotes methods with features of the average CNN activation of all objects above a detection confidence threshold.
DeepFE-rcnn \cite{karpathy2014fragment} uses a fragment mapping strategy to better exploit the object detection results.
The results show that using these features will improve the performance.
Even without the help from the object detection methods, however, our method performs better than these methods in most of the evaluation metrics.
We will develop our framework using better image features in the future work.

\begin{table}[htb]
	\centering
\begin{tabular}{l|cccc|cccc}
\hline
      & \multicolumn{4}{c|}{Sentence Retrival (Image to Text)} & \multicolumn{4}{c}{Image Retrival (Text to Image)} \\
\hline
      & R@1   & R@5   & R@10  & Med r & R@1   & R@5   & R@10  & Med r \\
\hline
Random & 0.1   & 0.5   & 1.0   & 631   & 0.1   & 0.5   & 1.0   & 500 \\
Socher-decaf \cite{socher2014grounded} & 4.5   & 18.0  & 28.6  & 32    & 6.1   & 18.5  & 29.0  & 29 \\
Socher-avg-rcnn \cite{socher2014grounded} & 6.0   & 22.7  & 34.0  & 23    & 6.6   & 21.6  & 31.7  & 25 \\
DeViSE-avg-rcnn \cite{frome2013devise} & 4.8   & 16.5  & 27.3  & 28    & 5.9   & 20.1  & 29.6  & 29 \\
DeepFE-decaf \cite{karpathy2014fragment} & 5.9   & 19.2  & 27.3  & 34    & 5.2   & 17.6  & 26.5  & 32 \\
DeepFE-rcnn \cite{karpathy2014fragment} & 12.6  & 32.9  & 44.0  & 14    & 9.7   & 29.6  & \textbf{42.5} & \textbf{15} \\
Ours-m-RNN-decaf & \textbf{14.5} & \textbf{37.2} & \textbf{48.5} & \textbf{11} & \textbf{11.5} & \textbf{31.0} & 42.4  & \textbf{15}\\
\hline
\end{tabular}%
	\caption{Results of R@K and median rank (Med r) for Flickr8K dataset. Note DeepFE-rcnn uses more sophisticated image features than we do}
	\label{tab:flickr8K_ret}
\end{table}

We report the results of generated sentences in Table \ref{tab:flickr8_30K_gen}.
There is no publicly available algorithm that reported results on this dataset.
So we compared our m-RNN model with the Ours-RNN-Base model.
The m-RNN model performs much better than this baseline both in terms of the perplexity and BLEU scores.

\subsection{Results on flickr30K}

This dataset is a new dataset and there are only a few methods report their retrieval results on it so far.
We first show the R@K evaluation metric in Table \ref{tab:flickr30K_ret}.
Our method outperforms the state-of-the-art methods in most of the evaluation metrics.
The results of the sentence generation task with a comparison of our RNN baseline are shown in Table \ref{tab:flickr8_30K_gen}.

\vspace{-0.5em}
\begin{table}[htb]
	\centering
\begin{tabular}{l|cccc|cccc}
\hline
      & \multicolumn{4}{c|}{Sentence Retrival (Image to Text)} & \multicolumn{4}{c}{Image Retrival (Text to Image)} \\
\hline
      & R@1   & R@5   & R@10  & Med r & R@1   & R@5   & R@10  & Med r \\
Random & 0.1   & 0.6   & 1.1   & 631   & 0.1   & 0.5   & 1.0   & 500 \\
DeViSE-avg-rcnn \cite{frome2013devise} & 4.8   & 16.5  & 27.3  & 28    & 5.9   & 20.1  & 29.6  & 29 \\
DeepFE-rcnn \cite{karpathy2014fragment} & 16.4  & \textbf{40.2} & \textbf{54.7} & \textbf{8} & 10.3  & \textbf{31.4} & \textbf{44.5} & \textbf{13} \\
Ours-m-RNN-decaf & \textbf{18.4} & \textbf{40.2} & 50.9  & 10    & \textbf{12.6} & 31.2  & 41.5  & 16 \\
\hline
\end{tabular}%
\vspace{-0.3em}
	\caption{Results of R@K and median rank (Med r) for Flickr30K dataset.}
	\label{tab:flickr30K_ret}
\end{table}

\begin{table}[htb]
	\centering
\begin{tabular}{l|rrrr|rrrr}
\hline
      & \multicolumn{4}{c|}{Flickr 8K} & \multicolumn{4}{c}{Flickr 30K} \\
\hline
      & $\mathcal{PPL}$  & B-1   & B-2   & B-3   & $\mathcal{PPL}$  & B-1   & B-2   & B-3 \\
\hline
Ours-RNN-Base & 30.39 & 0.4383 & 0.1849 & 0.1339 & 43.96 & 0.4699 & 0.1964 & 0.1252 \\
Ours-m-RNN & \textbf{24.39} & \textbf{0.5778} & \textbf{0.2751} & \textbf{0.2307} & \textbf{35.11} & \textbf{0.5479} & \textbf{0.2392} & \textbf{0.1952} \\
\hline
\end{tabular}%
\caption{Results of generated sentences in the Flickr8K and Flickr 30K dataset. }
\label{tab:flickr8_30K_gen}
\end{table}

\section{Conclusion}

We propose a multimodal Recurrent Neural Network (m-RNN) framework that performs at the state-of-the-art in three tasks: sentence generation, sentence retrieval given query image and image retrieval given query sentence.
Our m-RNN can be extended to use more complex image features (e.g. object detection features) and more sophisticated language models.

{\small
\bibliographystyle{ieee}
\bibliography{egbib}
}

\end{document}